\documentclass[conference]{IEEEtran}
\IEEEoverridecommandlockouts
\usepackage{cite}
\usepackage{amsmath,amssymb,amsfonts}
\usepackage{algorithmic}
\newcounter{algctr}
\usepackage{graphicx}
\usepackage{textcomp}
\usepackage{booktabs}
\usepackage{arydshln}
\usepackage{bm}

\def\BibTeX{{\rm B\kern-.05em{\sc i\kern-.025em b}\kern-.08em
    T\kern-.1667em\lower.7ex\hbox{E}\kern-.125emX}}
\begin{document}

\title{Linear Stability Analysis of an INDI Pitch-Rate Controller
       under Model Mismatch for a Tilt-Rotor VTOL UAV}

\author{%

\IEEEauthorblockN{Lorenzo Schenk}
\IEEEauthorblockA{\textit{IDSC}\\ ETH Zurich\\
Zurich, Switzerland\\
lschen@ethz.ch}
\and
\IEEEauthorblockN{Guillaume Ducard}
\IEEEauthorblockA{\textit{I3S Laboratory}\\ Univ. C\^{o}te d'Azur\\
Sophia Antipolis, France\\
guillaume.ducard@univ-cotedazur.fr}
\and
\IEEEauthorblockN{Christopher Onder}
\IEEEauthorblockA{\textit{IDSC}\\ ETH Zurich\\
Zurich, Switzerland\\
onder@idsc.mavt.ethz.ch}
\thanks{*This work was supported by the French government through the France 2030 investment plan managed by the National Research Agency (ANR), as part of the Initiative of Excellence Université Côte d’Azur under reference number ANR-15-IDEX-01.}
}

\maketitle

\begin{abstract}
Incremental Nonlinear Dynamic Inversion (INDI) is attractive for unmanned aerial vehicle (UAV) flight control because
it reduces dependence on a full aerodynamic model while retaining strong
disturbance-rejection capability. For a tilt-rotor vertical takeoff and landing (VTOL) architecture, however, the admissible
model-mismatch range of the fast inner loop is still not characterized analytically in a
parameter-explicit way. This paper isolates the pitch-rate/elevon subchannel of an existing
cascaded INDI controller and studies its linear stability under model mismatch. A closed-form
fifth-order transfer function is derived for the full
controller--estimator--actuator--plant interconnection, and stability is characterized through the
Routh--Hurwitz criterion over a parameterized linear model. Two representative three-parameter
sweeps produce interpretable stability regions. Based on these feasibility maps, two
uncertainty-aware tuning procedures are proposed: a robustness-oriented design that maximizes a
weighted worst-case combination of gain margin and phase margin, and a performance-oriented design that
maximizes worst-case closed-loop bandwidth subject to margin constraints. The results show that
actuator lag and inertia mismatch are comparatively benign at nominal gain, whereas
control-effectiveness mismatch, particularly a sign error in the allocation, is the most dangerous
destabilizing factor, leading to concrete tuning recommendations for conservative and aggressive
operating conditions.
\end{abstract}


\section{Introduction}

\subsection{INDI: Significance and Stability Challenges}

Incremental Nonlinear Dynamic Inversion (INDI) has emerged as a leading flight-control
approach for both rotary- and fixed-wing UAVs
\cite{schlatter2024indi,smeur2016adaptive,tal2022global}.
Unlike classical nonlinear dynamic inversion (NDI)
\cite{ducard2008ndi1,ducard2008ndi2,ducard2009fault},
which relies on a more complete model cancellation, INDI uses measured angular
acceleration together with a local control-effectiveness model to generate incremental
control commands~\cite{schlatter2024indi}. This substantially reduces dependence on a full
aerodynamic model and often improves disturbance rejection in practice
\cite{smeur2016adaptive}, but the closed-loop stability margin still depends on how
actuator, filtering, inertia, and effectiveness mismatch enter the implemented loop.
Quantifying the admissible mismatch range is therefore still necessary, especially across
wide flight envelopes.

Recent work has also broadened the field through a comprehensive survey of INDI
components, stability, and robustness \cite{steinert2025survey2} and through adaptive
INDI with guaranteed stability for aerial manipulators under varying inertia and
control-effectiveness conditions \cite{park2025adaptive}. Nevertheless, for aircraft
inner loops, analytical treatments still either rely on simplified single-lag plant models
\cite{smeur2016adaptive}, address related robustness questions without providing a
parameter-explicit Routh characterization of the fast pitch-rate inner loop
\cite{veld2018stability}, or assess stability numerically and empirically
\cite{wang2019indistability}. A closed-form mismatch-explicit Routh--Hurwitz analysis of
the pitch-rate loop of a tilt-rotor VTOL is still missing.

\subsection{Contributions}

This paper makes the following contributions:

\begin{enumerate}
    \item A closed-form, fifth-order transfer function is derived for the
    pitch-rate/elevon subchannel of the cascaded INDI controller~\cite{schlatter2024indi},
    explicitly accounting for actuator lag, estimator filtering, and inverse
    control allocation.
    \item The Routh--Hurwitz criterion is used to obtain interpretable stability conditions
    and to visualize representative three-parameter stability regions.
    \item Two uncertainty-aware tuning procedures are formulated on top of the
    Routh-feasible set: a robustness-oriented design maximizing a weighted
    worst-case combination of gain and phase margin, and a performance-oriented
    design maximizing worst-case closed-loop bandwidth subject to margin constraints.
\end{enumerate}

The analysis isolates the pitch-rate/elevon subchannel from the cascaded INDI inner loop
of~\cite{schlatter2024indi}. This channel is selected because the longitudinal transition maneuver considered here is
dominated by pitch dynamics: acceleration from hover to forward flight requires a sustained
pitch-down command that drives the elevon toward its deflection limits, making the pitch
channel the most likely to become unstable under effectiveness mismatch. The reduction to the standalone pitch-rate loop is exact at nominal gain
because, under the inherited controller architecture of~\cite{schlatter2024indi},
the attitude-control and control-allocation blocks produce identical elevon
commands at the nominal operating point; deviations from this equivalence
constitute part of the model mismatch studied here. The resulting linear stability certificate is therefore a local inner-loop result, not a full-envelope MIMO guarantee for aggressive coupled flight or saturated actuation.
The remainder of the paper is structured as follows: Section~\ref{sec:system_controller}
derives the linear model, Section~\ref{sec:stability_analysis} characterizes its stability,
and Section~\ref{sec:results} presents the optimization results and design guidelines.

\section{System and Controller}
\label{sec:system_controller}

\subsection{Aircraft Pitch Model}

Small-perturbation longitudinal dynamics about a trimmed flight condition reduce to the
single-axis pitch equation~\cite{ducard2009fault}:
\begin{equation}
    I_{yy,m}\,\dot{q} = M_m\,\delta - d_q\,q,
    \label{eq:pitch_dyn}
\end{equation}
where $q$ is the pitch rate, $\delta$ the elevon deflection, $M_m \equiv
\partial M/\partial\delta$ the pitch-moment effectiveness, $d_q > 0$ the aerodynamic
pitch-rate damping (stabilizing), and $I_{yy,m}$ the true moment of inertia. Defining
\begin{equation}
    K_q \triangleq \frac{M_m}{I_{yy,m}}, \qquad
    \omega_p \triangleq \frac{d_q}{I_{yy,m}},
    \label{eq:plane_params}
\end{equation}
the plant transfer function is $G_\text{plane}(s) = K_q/(s + \omega_p)$.
Actuator dynamics are modeled as a first-order lag system:
\begin{equation}
    G_\text{act}(s) = \frac{1}{1 + \tau_\text{act}\,s}.
    \label{eq:gact}
\end{equation}
In the stable-parameter-range analysis of Section~\ref{sec:stable_parameter_ranges},
$d_q = 0$ is used, giving the integrating plant
$G_\text{plane}(s) = K_q/s$. Over the physically meaningful parameter
range considered here and with the nominal sign convention preserved,
positive aerodynamic damping adds non-negative corrections to the
intermediate characteristic-polynomial coefficients (cf.\
Table~\ref{tab:coeffs_compare}) and relaxes, rather than tightens, the
observed Routh conditions. We therefore use $d_q=0$ as a conservative
baseline in the stability sweeps, whereas the performance optimization
in Section~IV-B reintroduces a non-zero $d_q$ explicitly.

\subsection{INDI Pitch-Rate Controller}

\begin{figure}[!t]
    \centering
    \includegraphics[width=\columnwidth]{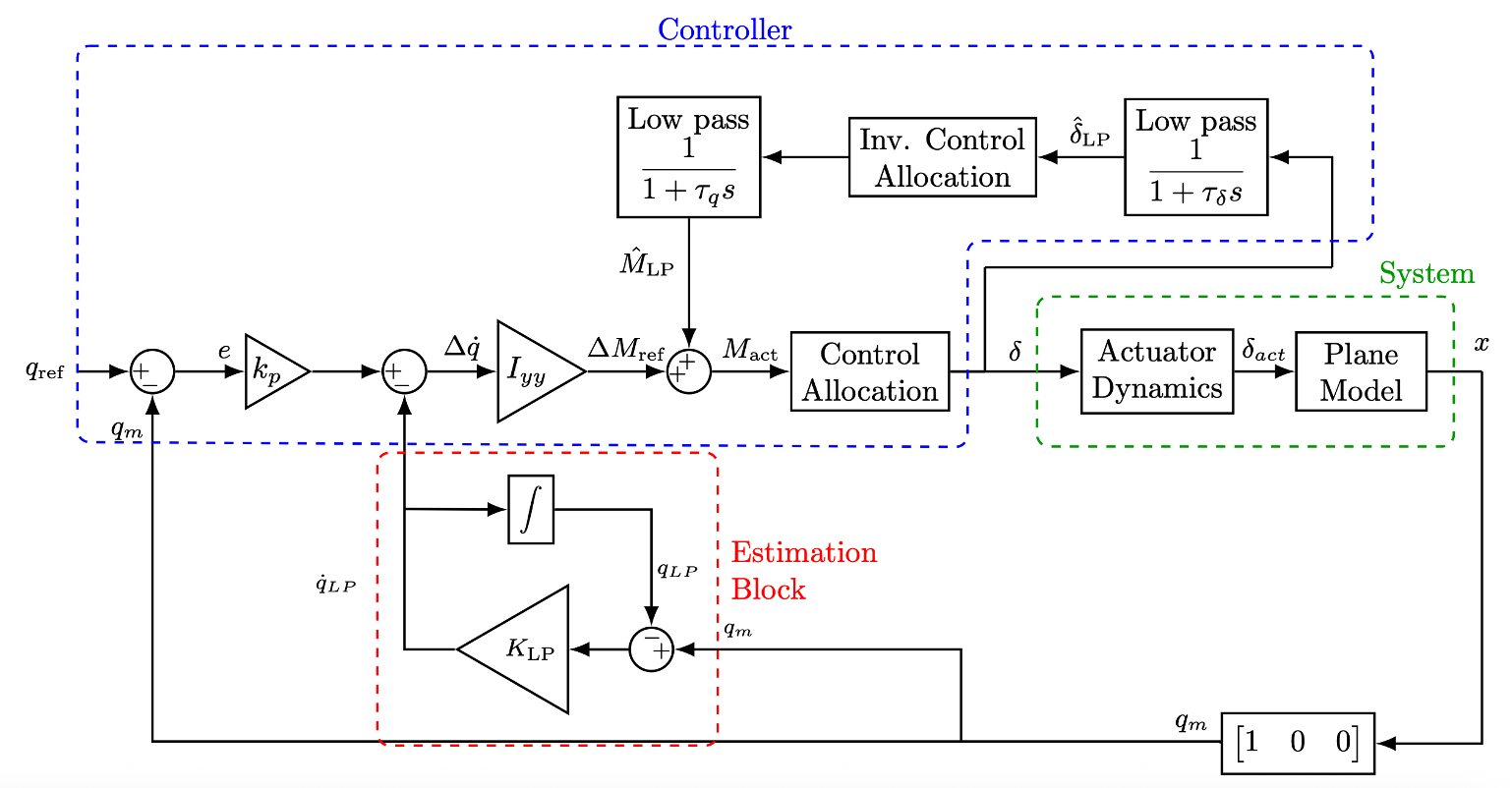}
    \caption{INDI pitch-rate loop derived from the inner-loop architecture
         of~\protect\cite{schlatter2024indi}. Blue: Controller; red: Estimation
         Block; green: System (actuator + plant).}
    \label{fig:controller}
\end{figure}

The reduced pitch-rate controller structure, shown in Fig.~\ref{fig:controller},
follows the inherited architecture of~\cite{schlatter2024indi} and consists of four
interconnected blocks: a proportional
error feedback, a pitch-rate estimator, an inverse control allocation, and a low-pass
filter chain.

\textbf{Proportional path.} The tracking error $e = q_\text{ref} - q_m$ is multiplied by the
proportional gain $k_p$. The rate-estimate branch subtracts the low-pass-filtered estimate
of $\dot{q}$, giving the incremental acceleration demand:
\begin{equation}
    \Delta\dot{q}(s) = k_p\,e(s) - \underbrace{\frac{k_\text{LP}\,s}{s + k_\text{LP}}\,q_m(s)}_{\dot q_{LP}(s)}.
    \label{eq:prop_path}
\end{equation}

\textbf{Moment reference.} The increment in pitch moment $\Delta M_{\mathrm{ref}}$ is obtained by multiplying $\Delta \dot{q}$ by the controller moment of inertia estimate $I_{yy,c}$.

\textbf{Moment Estimator (feedback compensation).} 
The estimated moment $\hat M_{LP}$ is reconstructed from the commanded elevon deflection $\delta$ through two cascaded first-order filters,
\begin{equation}
    F(s)=\frac{\hat M_{LP}(s)}{M_{act}(s)}=\frac{1}{(1+\tau_q s)(1+\tau_\delta s)} .
    \label{eq:F_est}
\end{equation}
In the matched implementation considered here, the same control-effectiveness
parameter $M_d$ is used in both the control allocation
$\Delta\delta=\Delta M_{\mathrm{ref}}/M_d$
and the estimator-side moment reconstruction. The internal estimator loop therefore reduces to
\begin{equation}
    B(s)=\frac{M_{\mathrm{act}}(s)}{\Delta M_{\mathrm{ref}}(s)}
    =\frac{1}{1-F(s)} .
    \label{eq:B_est}
\end{equation}
Model mismatch is introduced only through the difference between the assumed effectiveness $M_d$ and the true aircraft effectiveness $M_m$.

\textbf{Nominal time-alignment.} From Fig.~\ref{fig:controller}, the low-pass branch acting on the measured pitch rate $q_m$ is
\begin{equation*}
    \frac{k_\text{LP}}{s + k_\text{LP}} = \frac{1}{1 + \tau_q s},
    \qquad \tau_q \triangleq \frac{1}{k_\text{LP}},
\end{equation*}
which is aligned with the estimator filter when $k_\text{LP}=1/\tau_q$. In the same way, the reconstruction filter $1/(1+\tau_\delta s)$ matches the actuator dynamics $1/(1+\tau_\text{act}s)$ when $\tau_\delta=\tau_\text{act}$. These relations define the nominal time-alignment conditions
\begin{equation}
    k_\text{LP} = \frac{1}{\tau_q}, \qquad \tau_\delta = \tau_\text{act}.
    \label{eq:alignment}
\end{equation}
The quantities $k_{LP}$, $\tau_q$, and $\tau_\delta$ are design choices, whereas
$\tau_\text{act}$ is a fixed physical parameter. Satisfying \eqref{eq:alignment}
ensures that the reconstructed moment $\hat M_{LP}$ and angular-acceleration
estimate $\dot q_{LP}$ are based on the same delayed actuator signal; otherwise,
phase mismatch is introduced into the incremental cancellation path~\cite{veld2018stability}. Section~\ref{sec:stable_parameter_ranges} therefore varies $\tau_q$, $\tau_\delta$, and $k_\text{LP}$ independently to assess sensitivity to departures from nominal alignment.

\subsection{Closed-Loop Transfer Function}

The total effective plant seen by the proportional controller is
$P(s) = G_\text{plane}(s)\cdot G_\text{act}(s)\cdot (1/M_d)\cdot B(s)$.
Here \(P(s)\) is the forward map from the incremental moment reference to the measured pitch rate, i.e. \(q_m(s)=P(s)\Delta M_{\mathrm{ref}}(s)\). Since
\[
\Delta M_{\mathrm{ref}}(s)=I_{yy,c}\left[
k_p(q_{\mathrm{ref}}(s)-q_m(s))
-\frac{k_{LP}s}{s+k_{LP}}q_m(s)
\right],
\]
substitution gives
\[
q_m=P(s)I_{yy,c}k_pq_{\mathrm{ref}}
-P(s)I_{yy,c}\left(k_p+\frac{k_{LP}s}{s+k_{LP}}\right)q_m .
\]
Collecting the \(q_m\)-terms yields (8).
The closed-loop pitch-rate transfer function is
\begin{equation}
    \frac{q_m(s)}{q_\text{ref}(s)} =
    \frac{P(s)\,I_{yy,c}\,k_p}{1 + P(s)\,I_{yy,c}\!\left(k_p +
    \dfrac{k_\text{LP}\,s}{s+k_\text{LP}}\right)}.
    \label{eq:cl_tf}
\end{equation}

Expanding \eqref{eq:cl_tf} with $G_\text{plane}(s) = K_q/s$ yields a
fifth-order characteristic polynomial with coefficients that depend explicitly on
$\{k_p,\,k_\text{LP},\,I_{yy,c},\,\tau_\text{act},\,\tau_q,\,\tau_\delta,\,M_d,\,K_q\}$.

\subsection{Nominal Parameter Values}

Table~\ref{tab:nominal} collects the nominal values used throughout
the analysis. System parameters are fixed by the airframe and
operating point; controller parameters are design choices initialized
to match the physical plant.

\begin{table}[!t]
\caption{Nominal parameter values at the $V_a = 20\,\text{m/s}$
         fixed-wing trim point.}
\label{tab:nominal}
\renewcommand{\arraystretch}{1.12}
\centering
\begin{center}
\small
\begin{tabular}{llll}
\toprule
\textbf{Symbol} & \textbf{Value} & \textbf{Unit} & \textbf{Origin} \\
\midrule
\multicolumn{4}{l}{\textit{System parameters}} \\
$I_{yy,m}$      & 0.025  & kg\,m$^2$ & Physical airframe inertia \\
$\tau_\text{act}$ & 0.05 & s         & First-order servo model \\
$M_m$           & $\approx-8.4$ & N\,m/rad  & at $V_a{=}20\,\text{m/s} $ from \cite{schlatter2024indi} \\
$d_q$ & $\approx 3.92$ & N\,m\,s/rad &
  $-\tfrac{1}{4}\rho V_a^2 S\bar{c}^2 C_{mq}$~\cite{schlatter2024indi} \\
\midrule
\multicolumn{4}{l}{\textit{Controller parameters (nominal = matched)}} \\
$I_{yy,c}$      & $= I_{yy,m}$          & kg\,m$^2$ & Matched inertia \\
$M_d$           & $= M_m$               & N\,m/rad  & Matched effectiveness \\
$k_p$           & 20                    & rad/s     & Bandwidth target\cite{schlatter2024indi} \\
$\tau_q$        & 0.004                 & s         & Estimator corner \\
$k_\text{LP}$   & $= \frac{1}{\tau_q} = 250$    & rad/s     & Condition~\eqref{eq:alignment} \\
$\tau_\delta$   & $= \tau_\text{act} = 0.05$ & s    & Condition~\eqref{eq:alignment} \\
\bottomrule
\end{tabular}
\end{center}
\end{table}

The pitch-moment effectiveness $M_m = \bar{q}Sc\,C_{m\delta}$ is
computed from the dynamic pressure $\bar{q} = \tfrac{1}{2}\rho V_a^2$,
wing area $S$, chord $c$, and the pitch-moment derivative
$C_{m\delta}$~\cite{ducard2009fault}. At $V_a = 20\,\mathrm{m/s}$ this gives
$M_m \approx -8.4\,\mathrm{N\,m/rad}$ under the adopted NED nose-down
sign convention. This signed convention is used throughout, so
$K_q = M_m/I_{yy,m}$ and $M_d$ retain their physical sign; in the
matched nominal case $M_d$ and $M_m$ have the same sign, whereas a sign
error corresponds to $M_d$ crossing zero and reversing relative to
$M_m$. The aerodynamic pitch-rate damping $d_q \approx 3.92\,\text{N\,m\,s/rad}$ is computed
from the same trim condition. The controller
parameters $I_{yy,c}$ and $M_d$ are initialized to their physical
counterparts, and $k_\text{LP}$ and $\tau_\delta$ follow directly from
the time-alignment condition~\eqref{eq:alignment}. 
Throughout the stability analysis in \ref{sec:stable_parameter_ranges}, parameter perturbations are expressed
as percentages of their signed nominal values; for $M_d$ in particular,
$M_{d,\text{nom}} < 0$ under the adopted NED convention, so that a
perturbation crossing $0\%$ corresponds to an allocation-sign reversal,
identified in Section~\ref{sec:stable_parameter_ranges} as the most
dangerous destabilizing mode.

\section{Stability Analysis}
\label{sec:stability_analysis}

\subsection{Parameter Space and Model Mismatch}
The closed-loop characteristic polynomial depends on the system model  and controller parameters:
\[
    \underbrace{M_m, I_{yy,m}, \tau_{\mathrm{act}}, d_q}_\text{system parameters} \qquad
    \underbrace{M_d, k_p, I_{yy,c}, \tau_q, \tau_\delta, k_{LP}}_\text{controller parameters}
\]

The stability maps
consider two representative three-parameter subspaces: Sweep A varies
$\{M_d,\tau_{\mathrm{act}},k_p\}$ and Sweep B varies
$\{I_{yy,c},k_p,\tau_\delta\}$, while all remaining quantities are fixed at
their nominal values. In the optimization, the tunable controller vector is
\[
\theta=\{I_{yy,c},M_d,k_p,\tau_q,\tau_\delta,k_{LP}\},
\]
whereas plant uncertainty enters through multiplicative perturbations of the
physical parameters, as detailed in Section \ref{sec:opt}.

\begin{figure*}[!t]
    \centering
    \includegraphics[width=1\columnwidth]{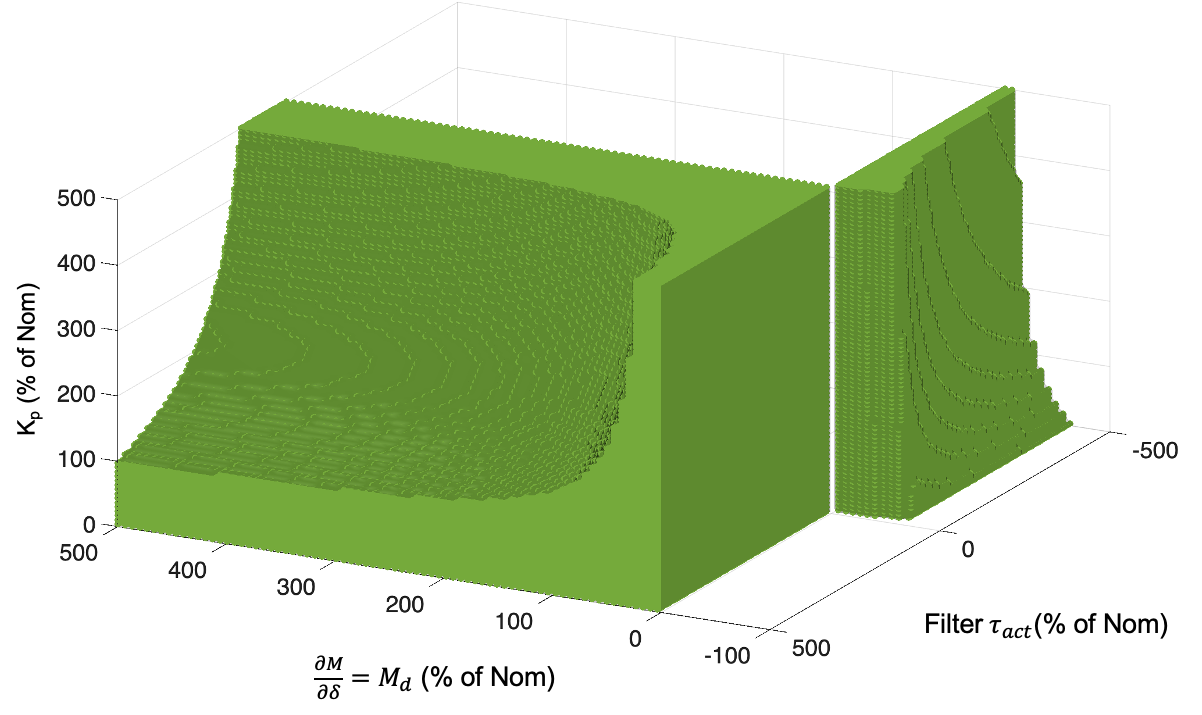}\hfill
    \includegraphics[width=1\columnwidth]{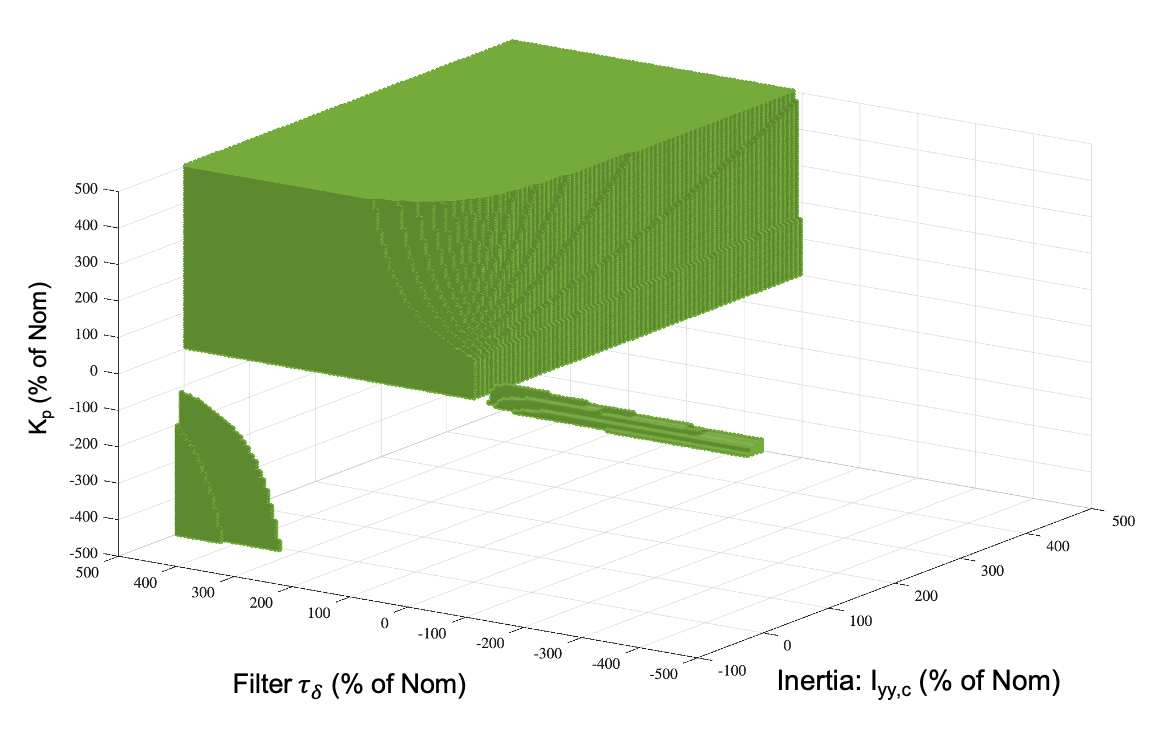}
    \caption{Stable parameter regions. Green voxels indicate parameter combinations for which all
        first-column entries of the Routh array are strictly positive, i.e.\
        the closed-loop system is asymptotically stable; the absence of voxels
        denotes instability.
        Left (\textbf{Sweep~A}): $\{M_d,\,\tau_\text{act},\,k_p\}$.
        Right (\textbf{Sweep~B}): $\{I_{yy,c},\,k_p,\,\tau_\delta\}$.
        Axes in percent of nominal; for $M_d$, percentages are relative to
        the signed nominal value $M_{d,\text{nom}}<0$; crossing $0\%$
        corresponds to an allocation-sign reversal.
        Each sweep contains a small disconnected stable island at negative $k_p$
        (visible in both panels)}
    \label{fig:stable_vol}
\end{figure*}

\subsection{Routh--Hurwitz Stability Criterion}

For a given parameter tuple $\bm{\theta}$, the characteristic polynomial of
\eqref{eq:cl_tf} is formed and the Routh array is evaluated. 

Table~\ref{tab:coeffs_compare} makes clear that positive aerodynamic damping enters as
additive corrections to the intermediate coefficients, which explains why the undamped
case is a conservative baseline, even though the full stability decision is still made
through the complete Routh array.

\begin{table}[h]
\centering
\caption{Characteristic polynomial coefficients: undamped baseline ($\omega_p=0$)
versus damped form. Here $G_{MM}=I_{yy,c}K_q/M_d$, $T_s=\tau_q+\tau_\delta$, and
$T_p=\tau_q\tau_\delta$. Positive aerodynamic damping adds only the listed
corrections $\delta C_i$, with $\delta C_5=\delta C_0=0$.}
\label{tab:coeffs_compare}
\scriptsize
\setlength{\tabcolsep}{4pt}
\renewcommand{\arraystretch}{1.2}
\begin{tabular}{c : p{0.34\columnwidth} : p{0.38\columnwidth}}
\toprule
Coeff. & Undamped $C_i^{(0)}$ ($\omega_p=0$) & Full $C_i = C_i^{(0)} + \delta C_i$ \\
\midrule
$C_5$ &
$\tau_\mathrm{act}T_p$ &
unchanged \\[2pt]

$C_4$ &
$\tau_\mathrm{act}T_s + T_p(1+\tau_\mathrm{act}k_{lp})$ &
$+\;\omega_p\tau_\mathrm{act}T_p$ \\[2pt]

$C_3$ &
$\tau_\mathrm{act}k_{lp}T_s + T_s + T_pk_{lp} + G_{MM}T_p(k_p+k_{lp})$ &
$+\;\omega_p[\tau_\mathrm{act}T_s + T_p(1+\tau_\mathrm{act}k_{lp})]$ \\[2pt]

$C_2$ &
$k_{lp}T_s + G_{MM}[k_pk_{lp}T_p+(k_p+k_{lp})T_s]$ &
$+\;\omega_p[T_s(1+\tau_\mathrm{act}k_{lp})+k_{lp}T_p]$ \\[2pt]

$C_1$ &
$G_{MM}[(k_p+k_{lp})+k_pk_{lp}T_s]$ &
$+\;\omega_p\,k_{lp}T_s$ \\[2pt]

$C_0$ &
$G_{MM}k_pk_{lp}$ &
unchanged \\
\bottomrule
\end{tabular}
\end{table}

To interpret the sampled Routh boundary, we inspected which first-column entry becomes
non-positive first along the numerically observed stability frontier \cite{franklin2006feedback}. Two recurring mechanisms dominate. First, the admissible proportional action is limited
by an effective loop-gain combination,
\[
\hat K \triangleq \frac{K_q I_{yy,c} k_p}{M_d},
\]
so smaller $|M_d|$, larger $k_p$, or larger $I_{yy,c}$ push the loop toward instability. Second,
the actuator and filter dynamics contribute through the corner frequencies
$\omega_a = 1/\tau_{act}$, $\omega_q = 1/\tau_q$, and $\omega_\delta = 1/\tau_\delta$, so
increasing any of these time constants reduces the admissible gain range. For compactness we
summarize these two recurring trends by the heuristic relations
\begin{align}
    \hat K &< f_1(\omega_a, \omega_q, \omega_\delta) \label{eq:condition_1} \\
    \omega_a \omega_q \omega_\delta &> f_2(\hat K) \label{eq:condition_2}
\end{align}
where $f_1$ and $f_2$ are induced by the corresponding Routh first-column expressions.
These relations are descriptive summaries of the sampled boundary, not reduced exact stability
tests; all numerical stability decisions in this paper are still made using the full
fifth-order Routh criterion.

Condition~\eqref{eq:condition_1} bounds the effective loop gain: underestimating $M_d$ inflates
$\hat{K}$ and drives the system toward instability. Condition~\eqref{eq:condition_2} requires the
product of corner frequencies to exceed a threshold set by $\hat{K}$: excessive lag from
the actuator, estimator, or alignment filter reduces this product and shrinks the stable
gain range, represented in the green volume.

\subsection{Stable Volume Visualization}
\label{sec:stable_volume_visualization}

Two complementary three-dimensional parameter sweeps are performed, shown in
Fig.~\ref{fig:stable_vol}. For each sweep, a dense Cartesian grid of $150^3$
points is evaluated over the normalized range $[-500\%,500\%]$ on each axis.
At every grid point, the full fifth-order characteristic polynomial is formed,
the complete Routh array is constructed, and the point is labeled stable if and
only if all first-column entries are strictly positive. The voxel plots
therefore represent sampled feasibility sets rather than fitted surfaces.

\textbf{Sweep A} varies $\{M_d,\tau_\mathrm{act},k_p\}$ and \textbf{Sweep B}
varies $\{I_{yy,c},k_p,\tau_\delta\}$, with all remaining parameters held at
their nominal values. A small disconnected stable region is also visible for
negative $k_p$ in both sweeps; this branch is noted for completeness but is not
pursued further because the present study focuses on practically relevant
positive-gain tuning.

\section{Results and Design Recommendations}
\label{sec:results}

\subsection{Stable Parameter Ranges}
\label{sec:stable_parameter_ranges}
\textbf{Sweep A} (Fig.~\ref{fig:stable_vol}, left) shows a stable
region whose admissible $k_p$ ceiling rises monotonically as $M_d$
decreases from 500\% of nominal toward 0\%, then vanishes abruptly
when $M_d$ crosses zero. Condition~\eqref{eq:condition_1} alone
would predict the ceiling should rise with $M_d$ (since larger $|M_d|$
reduces $\hat{K}$); the plot shows the opposite. The reason is that
$M_d$ enters the gain-loaded characteristic-polynomial terms through
$G_{MM} = I_{yy,c}K_q/M_d$. At large $M_d$, $G_{MM}$ is small,
which weakens $C_1$ and $C_3$ and makes the admissible $k_p$ ceiling
mainly set by the actuator--filter time scales rather than by $\hat{K}$
alone. As $M_d$ decreases, $G_{MM}$ grows, strengthening these terms
and raising the admissible ceiling. Larger $\tau_\text{act}$ uniformly
lowers this ceiling at all $M_d$, consistent with
condition~\eqref{eq:condition_2}: larger $\tau_\text{act}$ reduces
$\omega_a$ and tightens the corner-frequency product. Equivalently, the actuator then delivers the requested moment with excessive delay, so increasing \(k_p\) mainly increases phase lag rather than useful pitch-rate correction. When $M_d$
crosses zero, the allocation sign reverses and INDI becomes positive
feedback, which means that in practice the commanded incremental moment is applied with the wrong sign. No positive $k_p$ can stabilize this, and the entire
positive-gain stable region vanishes.

\textbf{Sweep B} (Fig.~\ref{fig:stable_vol}, right) shows a large
stable block across most of the physically meaningful region
($\tau_\delta > 0$, $I_{yy,c} > 0$), with the $k_p$ ceiling
decreasing as both $\tau_\delta$ and $I_{yy,c}$ simultaneously
approach 0\% of nominal. Both dominant conditions predict the
opposite: $I_{yy,c} \to 0^+$ reduces $\hat{K}$ (relaxing
condition~\eqref{eq:condition_1}) and $\tau_\delta \to 0^+$ increases
$\omega_\delta$ (relaxing condition~\eqref{eq:condition_2}). The
instability arises from a simultaneous collapse of the polynomial
structure: $\tau_\delta \to 0^+$ makes
$C_5 = \tau_\text{act}\tau_q\tau_\delta \to 0$, while
$I_{yy,c} \to 0^+$ drives $G_{MM} \to 0$ and weakens the
gain-carrying low-order terms $C_0$ and $C_1$. This boundary is
therefore captured only by the full Routh test, not by the two
heuristic summaries alone. Physically, the problem is the simultaneous loss of two stabilizing mechanisms: \(\tau_\delta\to0\) removes one reconstruction dynamic from the estimator path, while \(I_{yy,c}\to0\) removes the low-frequency corrective moment generated from the pitch-rate error. Note that $\tau_\delta \to 0$ also constitutes a severe violation of the time-alignment
condition~\eqref{eq:alignment}, confirming that operating far below
$\tau_\delta = \tau_\text{act}$ is not safe despite the apparent
relaxation of the heuristic lag condition.

\subsection{Uncertainty-Aware Tuning}
\label{sec:opt}

\textbf{Robustness optimization ($d_q = 0$, $\pm 25\%$).} The uncertainty set
$\mathcal{C}$ is defined by $\pm25\%$ perturbations of the three uncertain plant
parameters $\{\tau_\text{act},\,M_m,\,I_{yy,m}\}$, giving 8 vertex cases
$c\in\mathcal{C}$. The optimization is carried out on these extreme cases only,
so the resulting solution is the best one over the tested vertices, not a formal
optimum over the full interior of the uncertainty box. The objective is the
weighted worst-case combination of gain and phase margin:
\begin{equation}
    \max_{\bm{\theta}}\;\min_{c\in\mathcal{C}}\;
    \Bigl[0.5\,G_m(\bm{\theta},c)+0.5\,P_m(\bm{\theta},c)\Bigr].
    \label{eq:rob_opt}
\end{equation}
$G_m$ and $P_m$ are computed from the open-loop transfer function associated
with \eqref{eq:cl_tf}. Setting $d_q=0$ makes this study conservative with
respect to positive aerodynamic damping over the parameter range considered.

\indent \textbf{Performance optimization ($d_q > 0$, $\pm 10\%$).} Aerodynamic damping is
included as a fourth uncertain parameter, giving 16 corners $c \in \mathcal{C}$
($\pm 10\%$ on each of the four parameters). The objective is worst-case bandwidth
subject to margin constraints:
\begin{equation}
    \max_{\bm{\theta}}\;\min_{c\in\mathcal{C}}\;\omega_\text{BW}(\bm{\theta},c),
    \label{eq:perf_opt}
\end{equation}
\begin{equation*}
    \text{s.t.}\quad 45^\circ \leq P_m(\bm{\theta},c) \leq 60^\circ,\;
    G_m(\bm{\theta},c) \geq 6\,\text{dB},\; \forall c \in \mathcal{C}.
\end{equation*}

Here $\omega_\text{BW}$ denotes the first $-3\,\mathrm{dB}$ crossing of the
closed-loop transfer magnitude $|T(j\omega)|$, with
$T(s)=q_m(s)/q_\text{ref}(s)$. The 6~dB gain floor provides a standard $2\times$ gain safety factor.

Both problems are solved by exhaustive grid search over the Routh-feasible
controller grid
\[
\Theta = \Theta_{I_{yy,c}} \times \Theta_{M_d} \times \Theta_{k_p} \times
\Theta_{\tau_q} \times \Theta_{\tau_\delta} \times \Theta_{k_{LP}} .
\]
Algorithm~\ref{alg:opt} summarizes the search procedure. This grid is distinct
from the three-parameter visualization sweeps of
Section~\ref{sec:stable_volume_visualization}, which only display representative
slices of the stability boundary.

\begin{center}
\begin{minipage}{0.97\columnwidth}
\refstepcounter{algctr}\label{alg:opt}
\hrule
\vspace{3pt}
\footnotesize
\textbf{Algorithm \thealgctr: Uncertainty-Aware Controller Parameter Search}
\vspace{3pt}
\hrule
\begin{algorithmic}[1]
\REQUIRE Uncertainty set $\mathcal{C}$, parameter grid $\Theta$, objective
\ENSURE Optimal parameter set $\bm{\theta}^*$
\STATE Evaluate the Routh criterion for all $\bm{\theta}\in\Theta$; retain $\Theta_{\mathrm{stable}}$
\FOR{each $\bm{\theta}\in\Theta_{\mathrm{stable}}$}
    \STATE Compute Routh feasibility, \(G_m\), \(P_m\), and \(\omega_{BW}\) for all \(c\in\mathcal{C}\)
    \STATE Evaluate objective \eqref{eq:rob_opt} or \eqref{eq:perf_opt}
\ENDFOR
\STATE Return $\bm{\theta}^*=\arg\max_{\bm{\theta}\in\Theta_{\mathrm{stable}}}$ objective
\end{algorithmic}
\vspace{2pt}
\hrule
\end{minipage}
\end{center}
In the implementation, the Routh--Hurwitz test is used as a cheap pre-filter before gain-margin, phase-margin, and bandwidth evaluations are performed at the uncertainty corners. With \(N\) grid points per controller parameter and \(n_c\) uncertain plant parameters evaluated at box vertices, the uniform search scales as \(\mathcal{O}(N^6 \cdot 2^{n_c})\). For higher-dimensional problems, the same code structure could be made more scalable by replacing the uniform grid with coarse-to-fine refinement around Routh-feasible or Pareto-relevant regions.

\subsection{Recommended Parameter Sets}

\begin{figure*}[!t]
    \centering
    \includegraphics[width=0.94\columnwidth]{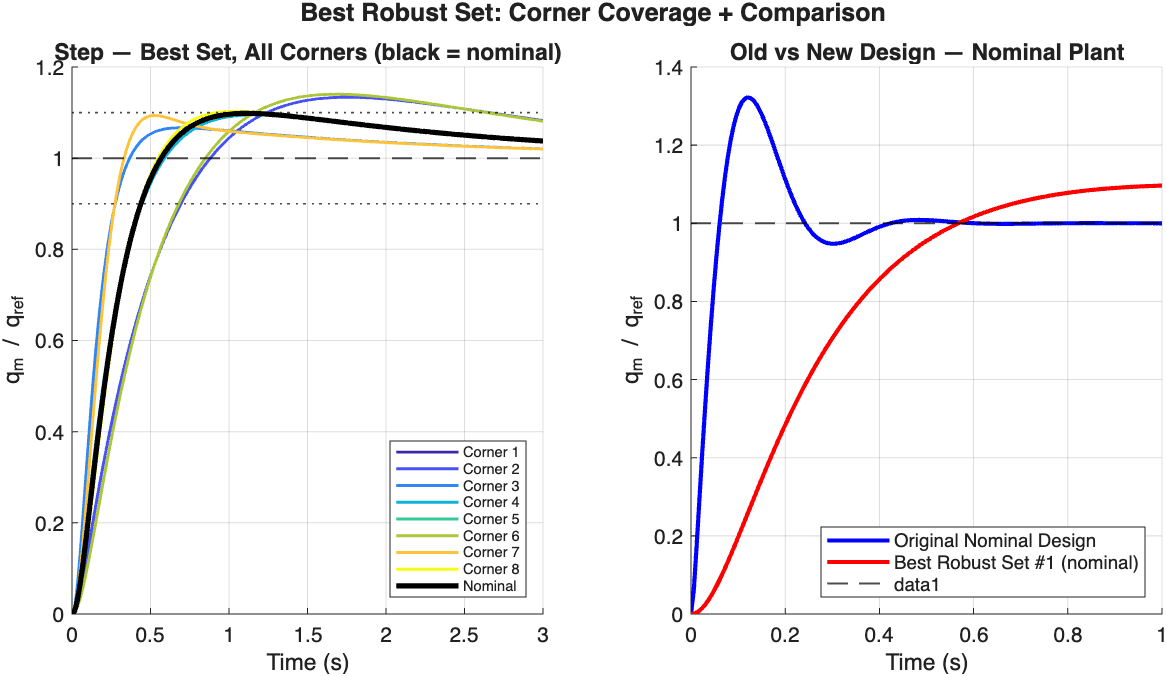}\hfill
    \includegraphics[width=0.94\columnwidth]{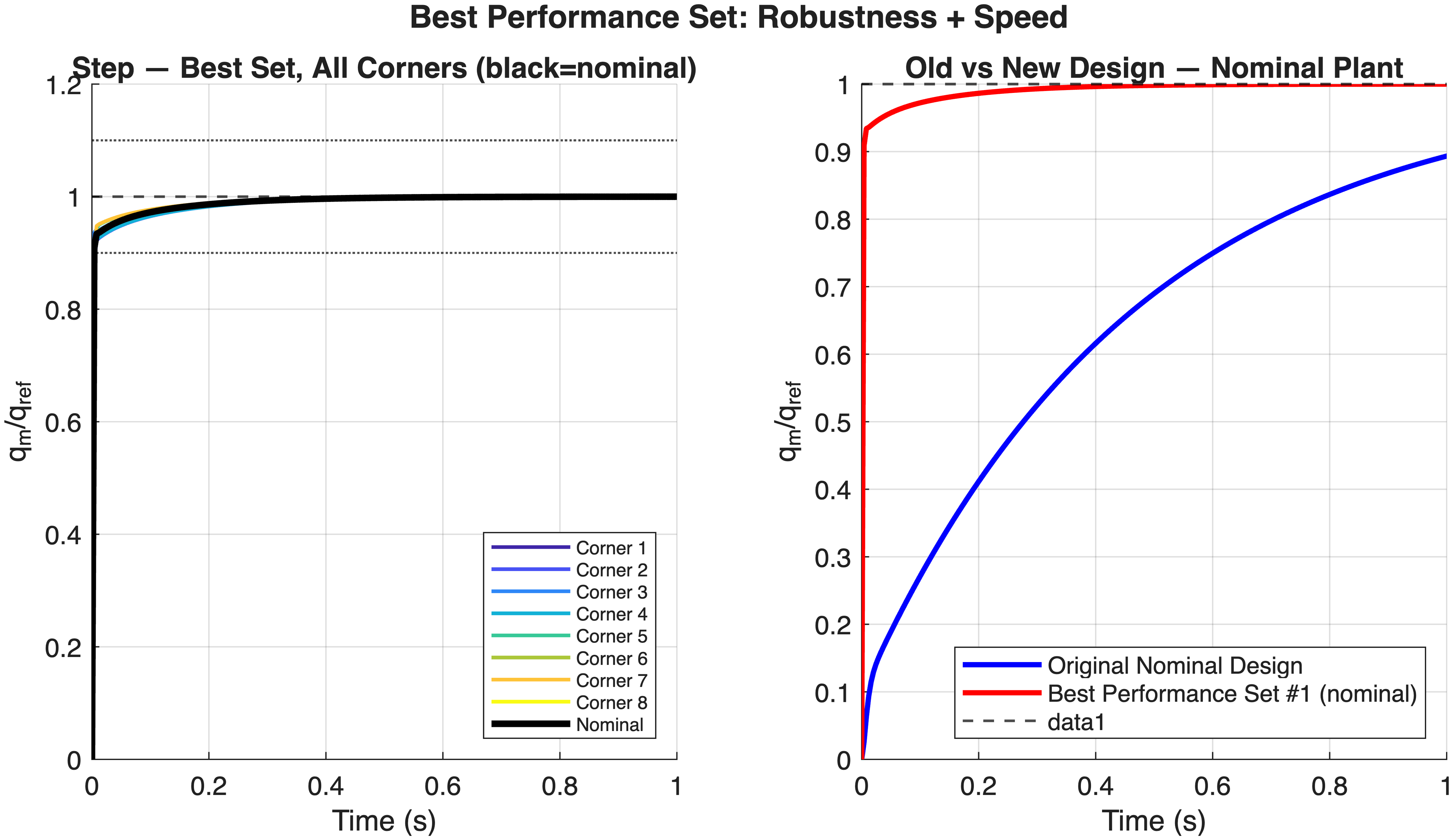}
    \caption{Step responses for the two optimal parameter sets. Black \& red $=$ nominal plant;
             colored lines $=$ all uncertainty corners. Left: robustness-optimal set
             ($\pm 25\%$, $d_q{=}0$): tight overdamped bundle, settling in
             $\approx 2.5\,\text{s}$, compared to the original design (right sub-panel).
             Right: performance-optimal set ($\pm 10\%$, $d_q{>}0$): all corners
             settle within $\approx 0.1\,\text{s}$, $\approx 4\times$ faster than the
             original design on the nominal plant.}
    \label{fig:opt_results}
\end{figure*}

The two optimal sets do not follow the same structural logic; rather, they reflect two
different operating objectives under two different uncertainty descriptions. The
robustness-optimal set drives the loop into a conservative regime by choosing a very small
$k_p$, fast filters, and a large $M_d$, thereby reducing the effective loop gain and producing
large stability margins. The performance-optimal set, by contrast, pushes several parameters
toward the edge of the admissible search box and should therefore be interpreted as a
boundary-seeking high-bandwidth candidate within the chosen grid, not as evidence of the same
tuning pattern. The clearest common feature between the two solutions is a small $\tau_q$,
which indicates that a fast rate-estimation path is beneficial in both formulations.

\begin{table}[h]
\caption{Optimized Parameter Sets Relative to Nominal.
\small $(\pm25\%,\,d_q{=}0)$: independent $\pm25\%$ variations of
$\{\tau_\text{act}, M_m, I_{yy,m}\}$, $d_q$ fixed at zero.
$(\pm10\%,\,d_q{>}0)$: $\pm10\%$ variations of all four parameters
including $d_q > 0$.}
\label{tab:params}
\centering
\renewcommand{\arraystretch}{1.12}
\begin{tabular}{|l|c|c|}
\hline
\textbf{Param.}
  & \textbf{Robustness} ($\pm 25\%$, $d_q{=}0$)
  & \textbf{Performance} ($\pm 10\%$, $d_q{>}0$) \\
\hline
$I_{yy,c}^{\,\text{opt}}$ & $0.05\cdot I_{yy,c}^{\,\text{nom}}$ & $4.29\cdot I_{yy,c}^{\,\text{nom}}$ \\
$k_\text{LP}^{\,\text{opt}}$ & $0.75\cdot k_\text{LP}^{\,\text{nom}}$ & $5.0\cdot k_\text{LP}^{\,\text{nom}}$ \\
$\tau_\delta^{\,\text{opt}}$ & $0.05\cdot\tau_\delta^{\,\text{nom}}$ & $2.88\cdot\tau_\delta^{\,\text{nom}}$ \\
$k_p^{\,\text{opt}}$ & $0.02\cdot k_p^{\,\text{nom}}$ & $1.45\cdot k_p^{\,\text{nom}}$ \\
$\tau_q^{\,\text{opt}}$ & $0.05\cdot\tau_q^{\,\text{nom}}$ & $0.05\cdot\tau_q^{\,\text{nom}}$ \\
$M_d^{\,\text{opt}}$ & $5.0\cdot M_d^{\,\text{nom}}$ & $0.05\cdot M_d^{\,\text{nom}}$ \\
\hline
\end{tabular}
\end{table}

The \textbf{robustness-optimal} set achieves guaranteed gain margins above $\approx
30\,\text{dB}$ and phase margins near $60^\circ$ across all eight corners, but sacrifices performance at the cost
of a slow overdamped step response (settling time $\approx 2.5\,\text{s}$) (see Fig.~\ref{fig:opt_results}, left), which makes it unsuitable as the nominal tuning for agile UAV operation. Its value lies instead in serving as a \emph{conservative stability
certificate}: a verified safe operating point that can be used as a
validated baseline when plant parameters are poorly characterized. In deployment, this set is therefore best interpreted as a low-confidence fallback tuning, for example during early flight testing or after suspected effectiveness degradation. Agility can then be recovered by moving toward the performance-oriented design when the uncertainty bounds are tighter and positive aerodynamic damping is sufficiently characterized.


The \textbf{performance-optimal} set achieves a substantially larger worst-case
closed-loop bandwidth than the nominal design while satisfying the
$[45^\circ, 60^\circ]$ phase-margin window and the $6\,\text{dB}$ gain floor.
In the current numerical sweep, the largest qualified bandwidth values are finite and occur
around $\omega_{BW}\approx 1.6\times 10^3\,\mathrm{rad/s}$. This high value indicates that, within the linear model and the imposed gain/phase-margin constraints, the modeled actuator lag, filtering, and mismatch are not the active bandwidth-limiting factors for this parameter combination. In practice, the limiting mechanisms would be hard elevon deflection and rate limits, sampling delay, sensor bandwidth, and structural constraints, which are not included in (\ref{eq:perf_opt}). The corresponding step responses in Fig.~\ref{fig:opt_results} are markedly faster than the nominal design, but this improvement requires tighter uncertainty bounds and knowledge of nominal $d_q$.


\begin{figure}[h]
    \centering
    \includegraphics[width=0.48\columnwidth]{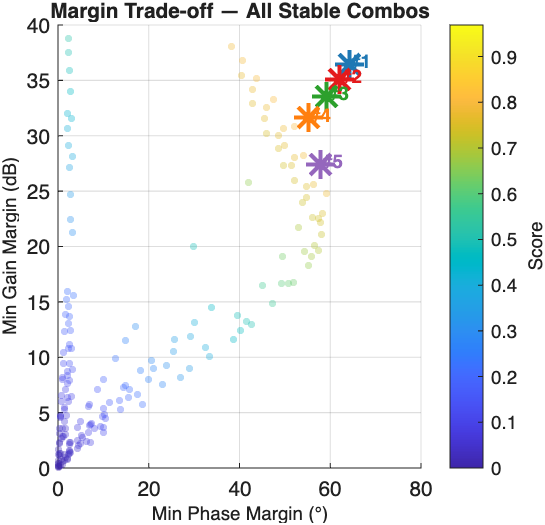}\hfill
    \includegraphics[width=0.48\columnwidth]{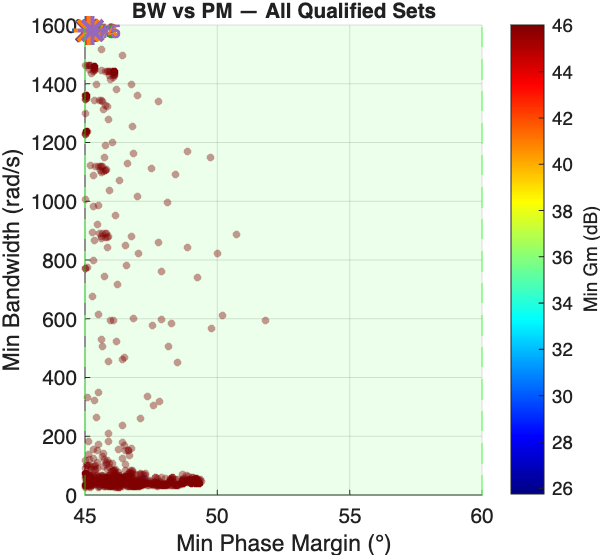}
    \caption{Left: worst-case gain/phase-margin trade-off over all Routh-feasible
             tunings for the robustness optimization~\eqref{eq:rob_opt}. Right: worst-case bandwidth versus worst-case phase margin for
             the performance optimization~\eqref{eq:perf_opt}. In both panels, the top-5 candidates are highlighted.}
    \label{fig:opt_solutions}
\end{figure}

\subsection{Design Guidelines}
The numerical study supports the following practical guidelines.

\begin{enumerate}
    \item \emph{Choose the control allocation gain $M_d$ and the controller gain $k_p$ jointly, and guarantee the correct sign of $M_d$.}
Sweep~A shows that the admissible $k_p$ ceiling increases as $M_d$ moves from large-magnitude values toward zero while preserving the nominal sign, but vanishes abruptly when $M_d$ crosses zero. A sign error in $M_d$ is therefore the most dangerous failure mode: it reverses the allocation feedback and makes stabilization with any positive $k_p$ impossible.

     \item \emph{Keep $\tau_\delta$ and $I_{yy,c}$ near their nominal values at
    nominal gain, or reduce $k_p$ accordingly.}
    Sweep B shows that simultaneously driving $\tau_\delta$ and $I_{yy,c}$
    toward 0\% of nominal destabilizes the loop by collapsing the
    fifth-order polynomial structure: $\tau_\delta \to 0$ removes the
    highest-order dynamic term $C_5 = \tau_\text{act}\tau_q\tau_\delta$,
    while $I_{yy,c} \to 0$ weakens the low-order gain-carrying terms
    through $G_{MM} \to 0$. At very small $k_p$ this mechanism is much less
    severe, which is why the robustness-optimal set can combine small
    $\tau_\delta$ and $I_{yy,c}$ with $k_p \approx 0$. At nominal or high
    $k_p$, the matched values $\tau_\delta = \tau_\text{act}$ and
    $I_{yy,c} = I_{yy,m}$ should be respected.
        
    \item Use the $d_q=0$ design as a conservative linear baseline when
    aerodynamic damping is uncertain. A tuning that is stable at
    $d_q=0$ is therefore an appropriate conservative starting point when
    the true damping is not known accurately.

    \item \emph{Validate the performance-optimal solution within its uncertainty bounds.}
    The performance-optimal solution in Section~\ref{sec:opt} is derived under
    $d_q > 0$ and therefore requires knowledge of the \emph{nominal}
    value of $d_q$, not its exact value; the $\pm10\%$
    uncertainty margin on $d_q$ demonstrates that moderate uncertainty
    in the damping estimate is tolerable. The step responses in Fig.~\ref{fig:opt_results}
    confirm the improved bandwidth for all evaluated corners, but the result should not
    be extrapolated beyond the tested uncertainty range or to operating points where
    $d_q$ is poorly characterized.
\end{enumerate}

\section{Conclusion}

This paper presented a systematic linear stability analysis of an INDI pitch-rate
controller for a tilt-rotor VTOL UAV under model mismatch. A closed-form fifth-order transfer
function was derived and the Routh--Hurwitz criterion was used to characterize stability
regions in two three-parameter subspaces. The stability volumes reveal that actuator lag and inertia mismatch are less destabilizing than control-effectiveness mismatch over the displayed sweep ranges: large $M_d$ combined with large $k_p$ collapses the gain-dependent Routh
coefficients, and a sign error in $M_d$ eliminates all positive-gain stability abruptly. Two uncertainty-aware tuning procedures were formulated
and solved, providing concrete recommendations for robustness-priority and
performance-priority conditions together with a quantification of the resulting
speed--robustness trade-off.

Within the parameter range considered here, the $d_q = 0$ baseline is
conservative: positive aerodynamic damping shifts the coefficient set in
a stabilizing direction and did not tighten the observed stability
boundary. Large-amplitude maneuvers, trim-point shifts during transition, and actuator saturation introduce nonlinear effects that can change effective gain and phase lag, thereby tightening or shifting the Routh boundary derived here. Future work will extend the stability
certificate to this nonlinear saturated regime via Sum-of-Squares Lyapunov methods \cite{detailleur2025SOS},
enabling certified stability regions that account for actuator saturation directly.
In addition, while the pitch focus is motivated by the longitudinal transition mechanism discussed in Section~I, a more complete full-envelope certificate should include roll/yaw coupling for asymmetric actuation, lateral disturbances, and non-longitudinal maneuvers.



\end{document}